\documentclass{article}

\usepackage[final,nonatbib]{neurips_2019}

\usepackage{etoolbox}
\makeatletter
\patchcmd{\maketitle}
 {\def\@makefnmark}
 {\def\@makefnmark{}\def\useless@macro}
 {}{}
\makeatother
\usepackage[utf8]{inputenc} % allow utf-8 input
\usepackage[T1]{fontenc}    % use 8-bit T1 fonts
\usepackage{hyperref}       % hyperlinks
\usepackage{url}            % simple URL typesetting
\usepackage{booktabs}       % professional-quality tables
\usepackage{amsfonts}       % blackboard math symbols
\usepackage{nicefrac}       % compact symbols for 1/2, etc.
\usepackage{microtype}      % microtypography
\usepackage{multirow}       % For multiple row
\usepackage{makecell}       % make a cell
\usepackage{color}
\usepackage{wrapfig}        % To wrap text around a figure
\usepackage{graphicx}
\usepackage{subcaption}
\usepackage{algorithm}
\usepackage{algorithmic}

\usepackage{amsmath,amsfonts,bm}

\def\vo{{\bm{o}}}
\def\vp{{\bm{p}}}

\def\vz{{\bm{z}}}

\def\mA{{\bm{A}}}
\def\mB{{\bm{B}}}

\def\mO{{\bm{O}}}

\def\mW{{\bm{W}}}
\def\mX{{\bm{X}}}

\def\gA{{\mathcal{A}}}

\def\gL{{\mathcal{L}}}

\def\gT{{\mathcal{T}}}
\def\gU{{\mathcal{U}}}

\def\sA{{\mathbb{A}}}

\def\sC{{\mathbb{C}}}
\def\sD{{\mathbb{D}}}
\def\sE{{\mathbb{E}}}

\def\sR{{\mathbb{R}}}

\def\sI{{\mathbb{I}}}

\def\0{\mathbf{0}}
\def\1{\mathbf{1}}

\def\Figref#1{Fig.~\ref{#1}}
\def\Secref#1{Sec.~\ref{#1}}
\def\Algref#1{Alg.~\ref{#1}}
\def\Tabref#1{Table~\ref{#1}}
\def\Eqref#1{Eq.~\eqref{#1}}
\def\NAME{{TAS}}

\title{Network Pruning via\\Transformable Architecture Search}

\author{
  Xuanyi Dong$^{\dagger\ddagger}$\thanks{This work was done when Xuanyi Dong was a research intern at Baidu Research.}~, Yi Yang$^{\dagger}$ \\
  $^{\dagger}$The ReLER Lab, University of Technology Sydney, $^{\ddagger}$Baidu Research\\
  \texttt{xuanyi.dong@student.uts.edu.au; yi.yang@uts.edu.au}
}

\begin{document}

\maketitle

\begin{abstract}
Network pruning reduces the computation costs of an over-parameterized network without performance damage. 
Prevailing pruning algorithms pre-define the width and depth of the pruned networks, and then transfer parameters from the unpruned network to pruned networks.
To break the structure limitation of the pruned networks, we propose to apply neural architecture search to search directly for a network with flexible channel and layer sizes.
The number of the channels/layers is learned by minimizing the loss of the pruned networks.
The feature map of the pruned network is an aggregation of K feature map fragments (generated by K networks of different sizes), which are sampled based on the probability distribution.
%The feature map of the pruned network is an aggregation of K feature map fragments. Fragments are generated by K networks of different sizes and sampled using a probability distribution.
% , whose channels/layers size is sampled from parameterized distributions.
%To fully train the distributions, the feature map is a weighted sum of K feature map fragments with K sampled from respective distribution.
% The feature map of the pruned networks during searching is an aggregation of K feature map fragments sampled based on the probability distribution.
The loss can be back-propagated not only to the network weights, but also to the parameterized distribution to explicitly tune the size of the channels/layers.
%To optimize these distributions, the feature map is an aggregation of K feature map fragments sampled according to respective distribution.
Specifically, we apply channel-wise interpolation to keep the feature map with different channel sizes aligned in the aggregation procedure.
The maximum probability for the size in each distribution serves as the width and depth of the pruned network, whose parameters are learned by knowledge transfer, e.g., knowledge distillation, from the original networks.
Experiments on CIFAR-10, CIFAR-100 and ImageNet demonstrate the effectiveness of our new perspective of network pruning compared to traditional network pruning algorithms.
Various searching and knowledge transfer approaches are conducted to show the effectiveness of the two components.
Code is at: \url{https://github.com/D-X-Y/NAS-Projects}.
\end{abstract}

\section{Introduction}\label{sec:intro}

Deep convolutional neural networks (CNNs) have become wider and deeper to achieve high performance on different applications~\cite{he2016deep,huang2017densely,zoph2017neural}.
Despite their great success, it is impractical to deploy them to resource constrained devices, such as mobile devices and drones.
\begin{wrapfigure}{r}{0.5\textwidth}
\vspace{-5mm}
  \begin{center}
    \includegraphics[width=0.5\textwidth]{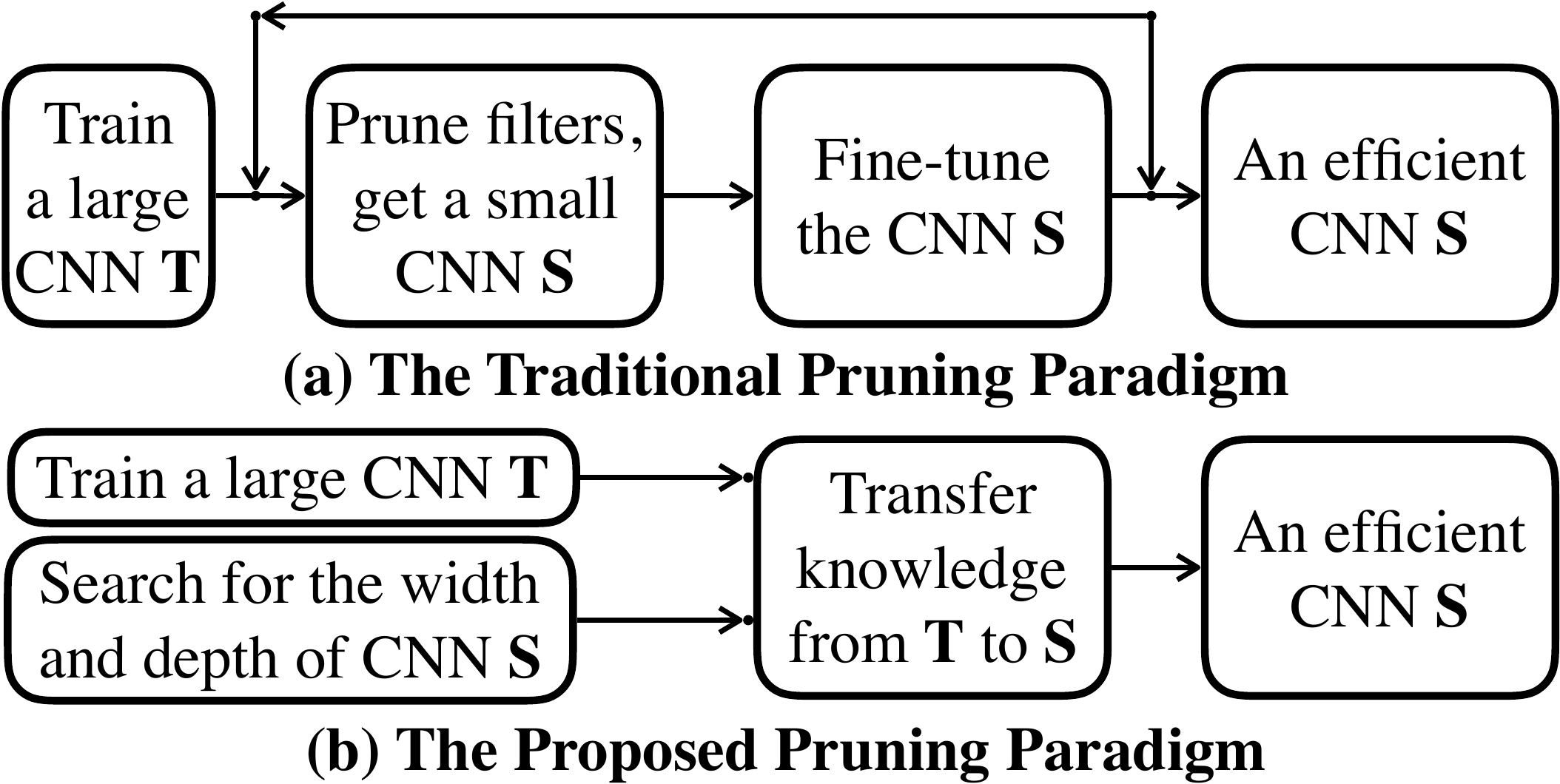}
  \end{center}
\vspace{-3mm}
\caption{A comparison between the typical pruning paradigm and the proposed paradigm.
}
\vspace{-5mm}
\label{fig:typical-vs-ours}
\end{wrapfigure}
A straightforward solution to address this problem is using network pruning~\cite{lecun1990optimal,han2016deep,han2016learning,he2017channel,he2018soft} to reduce the computation cost of over-parameterized CNNs. 
A typical pipeline for network pruning, as indicated in \Figref{fig:typical-vs-ours}(a), is achieved by removing the redundant filters and then fine-tuning the slashed networks, based on the original networks.
Different criteria for the importance of the filters are applied, such as L2-norm of the filter~\cite{li2017pruning}, reconstruction error~\cite{he2017channel}, and learnable scaling factor~\cite{liu2017learning}.
Lastly, researchers apply various fine-tuning strategies~\cite{li2017pruning,he2018soft} for the pruned network to efficiently transfer the parameters of the unpruned networks and maximize the performance of the pruned networks.

%Recent studies on network pruning lay in two modalities: unstructured pruning~\cite{lecun1990optimal,figurnov2017spatially,dong2017more,han2016deep} and structure pruning~\cite{li2017pruning,he2017channel,he2018soft,liu2019rethinking}.
%Unstructured pruning algorithms usually enforce the convolutional weights~\cite{lecun1990optimal} or feature maps~\cite{dong2017more} to be sparse, which can obtain a high theoretical acceleration and compression.
%However, since the non-zero weight entity of these methods is irregular, dedicated hardware and libraries are required~\cite{han2016eie}.
% , especially channel (filter) pruning,
% Structured pruning algorithms have attracted much attention~\cite{li2017pruning,ye2018rethinking,liu2019rethinking}. They can directly prune the convolutional filters~\cite{ye2018rethinking,liu2019rethinking} or the whole layer~\cite{figurnov2017spatially}. As a results, their pruned CNN can be easily developed with negligible extra effort.
%due to its simplicity and efficiency. As a results, their pruned CNN can be easily developed with negligible extra effort.
%In this work, we focus on structured network pruning.
% The traditional approaches first train a large CNN, prune some filters, and then fine-tune the pruned CNN~\cite{li2017pruning} (the top of \Figref{fig:typical-vs-ours}).
% A majority of researchers studied the criteria of selecting useless filters to prune, such as L2-norm~\cite{li2017pruning}, reconstruction error~\cite{he2017channel}, and learnable scaling factor~\cite{liu2017learning}.

Traditional network pruning approaches achieve effective impacts on network compacting while maintaining accuracy.
Their network structure is intuitively designed, e.g., pruning 30\% filters in each layer~\cite{li2017pruning,he2018soft}, predicting sparsity ratio~\cite{he2018amc} or leveraging regularization~\cite{alvarez2016learning}.
The accuracy of the pruned network is upper bounded by the hand-crafted structures or rules for structures.
To break this limitation, we apply Neural Architecture Search (NAS) to turn the design of the architecture structure into a learning procedure and propose a new paradigm for network pruning as explained in \Figref{fig:typical-vs-ours}(b).

% Most of these methods pre-define the structure of the pruned CNN, e.g., pruning 30\% filters in each layer~\cite{li2017pruning,he2018soft}.
% As a result, they suffer from a critical limitation: the pre-defined structure upper bound the accuracy of their pruning algorithms.
% Even if some methods allow different pruning ratio at each layer by predicting sparsity ratio~\cite{he2018amc} or using regularization~\cite{alvarez2016learning}, they still rely on heuristic filter selection criteria and does not directly optimize to find a robust structure.
% In conclusion, previous pruning methods suffer from the structure limitation.

%Motivated by the recent studies on neural architecture search (NAS)

%Prevailing NAS methods~\cite{liu2019darts,zoph2017neural,dong2019search,cai2018proxylessnas,real2019regularized} optimize the network topology while the focus in this literature of network pruning is the selection of network size.
Prevailing NAS methods~\cite{liu2019darts,zoph2017neural,dong2019search,cai2018proxylessnas,real2019regularized} optimize the network topology, while the focus of this paper is automated network size.
In order to satisfy the requirements and make a fair comparison between the previous pruning strategies, we propose a new NAS scheme termed Transformable Architecture Search ({\NAME}).
{\NAME} aims to search for the best size of a network instead of topology, regularized by minimization of the computation cost, e.g., floating point operations (FLOPs).
The parameters of the searched/pruned networks are then learned by knowledge transfer~\cite{hinton2014distilling,yim2017gift,zagoruyko2017paying}.
% knowledge distillaztion (KD)~\cite{hinton2014distilling}.
%
%
% In the proposed new paradigm, to prune a CNN $\mT$, we first search for a small CNN $\mS$ with flexible channel and layer sizes, and then learn the parameters of this searched $\mS$ by transferring from $\mT$ using knowledge distillation (KD)~\cite{hinton2014distilling}. The channel and layer sizes of $\mS$ are automatically learned to (1) maximize the accuracy of $\mS$ on a target dataset while (2) constrain the computational cost (e.g., FLOP) of $\mS$ being less than some pre-defined limitation.
% A straightforward idea of finding such channel and layer sizes is to use existing NAS approaches.
% However, since prevailing methods~\cite{liu2019darts,zoph2017neural,dong2019search,cai2018proxylessnas} focus on finding the network topology, they can not be readily applied in our scenario.

{\NAME} is a differentiable searching algorithm, which can search for the width and depth of the networks effectively and efficiently. Specifically, different candidates of channels/layers are attached with a learnable probability. The probability distribution is learned by back-propagating the loss generated by the pruned networks, whose feature map is an aggregation of K feature map fragments (outputs of networks in different sizes) sampled based on the probability distribution.
These feature maps of different channel sizes are aggregated with the help of channel-wise interpolation.
The maximum probability for the size in each distribution serves as the width and depth of the pruned network.

In experiments, we show that the searched architecture with parameters transferred by knowledge distillation (KD) outperforms previous state-of-the-art pruning methods on CIFAR-10, CIFAR-100 and ImageNet.
We also test different knowledge transfer approaches on architectures generated by traditional hand-crafted pruning approaches~\cite{li2017pruning,he2018soft} and random architecture search approach~\cite{liu2019darts}.
Consistent improvements on different architectures demonstrate the generality of knowledge transfer.

\section{Related Studies}\label{sec:relate}

%\subsection{Network pruning}\label{sec:relate-prune}
Network pruning~\cite{lecun1990optimal,liu2019rethinking} is an effective technique to compress and accelerate CNNs, and thus allows us to deploy efficient networks on hardware devices with limited storage and computation resources.
A variety of techniques have been proposed, such as low-rank decomposition~\cite{zhang2016accelerating}, weight pruning~\cite{hassibi1993second,lecun1990optimal,han2016learning,han2016deep}, channel pruning~\cite{he2018soft,liu2019rethinking}, dynamic computation~\cite{figurnov2017spatially,dong2017more} and quantization~\cite{hubara2017quantized,alizadeh2019empirical}.
They lie in two modalities: unstructured pruning~\cite{lecun1990optimal,figurnov2017spatially,dong2017more,han2016deep} and structured pruning~\cite{li2017pruning,he2017channel,he2018soft,liu2019rethinking}.
%rastegari2016xnor
%These algorithms lie in two modalities: unstructured pruning~\cite{lecun1990optimal,figurnov2017spatially,dong2017more,han2016deep} and structure pruning~\cite{li2017pruning,he2017channel,he2018soft,liu2019rethinking}.

%Some of above mentioned techniques can be categorized into
\textit{\textbf{Unstructured}} pruning methods~\cite{lecun1990optimal,figurnov2017spatially,dong2017more,han2016deep} usually enforce the convolutional weights~\cite{lecun1990optimal,hassibi1993second} or feature maps~\cite{dong2017more,figurnov2017spatially} to be sparse.
%, such as weight pruning and quantization.
The pioneers of unstructured pruning, LeCun~et~al.~\cite{lecun1990optimal} and Hassibi~et~al.~\cite{hassibi1993second}, investigated the use of the second-derivative information to prune weights of shallow CNNs.
After deep network was born in 2012~\cite{krizhevsky2012imagenet}, Han~et~al.~\cite{han2016deep,han2016learning,han2016eie} proposed a series of works to obtain highly compressed deep CNNs based on L2 regularization.
%Molchanov~et~al.~\cite{molchanov2017variational} explored variational dropout to increase the sparsity of networks.
After this development, many researchers explored different regularization techniques to improve the sparsity while preserve the accuracy, such as L0 regularization~\cite{louizos2018learning} and output sensitivity~\cite{tartaglione2018learning}.
%Louizos~et~al.~\cite{louizos2018learning} investigated the use of L0 regularization and relaxed the discrete nature of the L0  penalty for efficient continuous optimization.
%Tartaglione~et~al.~\cite{tartaglione2018learning} quantified the output sensitivity to each parameter as a kind of regularization.
%Others~\cite{dong2017more,figurnov2017spatially} studied dynamical computation for deep neural networks.
%Dong~et~al.~\cite{dong2017more} and Figurnov~et~al.~\cite{figurnov2017spatially} adaptively decided some parts of computation in the network inference procedure to be skipped for each example.
%These unstructured pruning methods can obtain a high theoretical acceleration and compression, while require dedicated hardware and libraries~\cite{han2016eie}.
Since these unstructured methods make a big network sparse instead of changing the whole structure of the network, they need dedicated design for dependencies~\cite{han2016eie} and specific hardware to speedup the inference procedure.
% These methods usually heavily rely on hardware and libraries~\cite{han2016eie}, they need additional efforts to achieve a good realistic speedup.

\textit{\textbf{Structured}} pruning methods~\cite{li2017pruning,he2017channel,he2018soft,liu2019rethinking} target the pruning of convolutional filters or whole layers, and thus the pruned networks can be easily developed and applied.
Early works in this field~\cite{alvarez2016learning,wen2016learning} leveraged a group Lasso to enable structured sparsity of deep networks.
% Alvarez~et~al.~\cite{alvarez2016learning} and Wen~et~al.~\cite{wen2016learning} leveraged the group Lasso to enable structured sparsity of deep networks.
%He~et~al.~\cite{he2017channel} proposed a LASSO regression based channel selection and least square reconstruction for channel pruning.
After that, Li~et~al.~\cite{li2017pruning} proposed the typical three-stage pruning paradigm (training a large network, pruning, re-training).
These pruning algorithms regard filters with a small norm as unimportant and tend to prune them, but this assumption does not hold in deep nonlinear networks~\cite{ye2018rethinking}. Therefore, many researchers focus on better criterion for the informative filters.
For example, Liu~et~al.~\cite{liu2017learning} leveraged a L1 regularization; Ye~et~al.~\cite{ye2018rethinking} applied a ISTA penalty; and He~et~al.~\cite{he2019pruning} utilized a geometric median-based criterion.
%For example, Liu~et~al.~\cite{liu2017learning} leveraged the scaling factor in Batch Normalization (BN)~\cite{ioffe2015batch} as an importance indicator for the corresponding convolutional filter with a simple L1 regularization.
%Ye~et~al.~\cite{ye2018rethinking} took a deep analysis into the smaller-norm-less-informative assumption and proposed to use the ISTA~\cite{beck2009fast} penalty for the BN scaling factor.
In contrast to previous pruning pipelines, our approach allows the number of channels/layers to be explicitly optimized so that the learned structure has high-performance and low-cost.\looseness-1
%we propose a new pruning paradigm: searching (the depth and width of a small network with desired FLOP), train (a large network), and transfer (the knowledge from the large network to the small network). This paradigm shows its simplicity and effectiveness in experiments.

Besides the criteria for informative filters, the importance of network structure was suggested in~\cite{liu2019rethinking}.
% Our work is related to some recent studies on pruning.
% From the perspective of NAS,
% Liu~et~al.~\cite{liu2019rethinking} suggested that the architecture of the pruned CNN is crucial for the accuracy. 
Some methods implicitly find a data-specific architecture~\cite{wen2016learning,alvarez2016learning,he2018amc}, by automatically determining the pruning and compression ratio of each layer.
%In contrast, we learn the architecture in the scenario of NAS, which explicitly optimizes the architecture.
In contrast, we explicitly discover the architecture using NAS.
Most previous NAS algorithms~\cite{zoph2017neural,dong2019search,liu2019darts,real2019regularized} automatically discover the topology structure of a neural network, while we focus on searching for the depth and width of a neural network.
Reinforcement learning (RL)-based~\cite{zoph2017neural,cai2018efficient} methods or evolutionary algorithm-based~\cite{real2019regularized} methods are possible to search networks with flexible width and depth, however, they require huge computational resources and cannot be directly used on large-scale target datasets.
Differentiable methods~\cite{dong2019search,liu2019darts,cai2018proxylessnas} dramatically decrease the computation costs but they usually assume that the number of channels in different searching candidates is the same.
{\NAME} is a differentiable NAS method, which is able to efficiently search for a transformable networks with flexible width and depth.
% This assumption does not hold in our scenario.
% {\NAME} applies channel-wise interpolation to keep networks with different channel sizes aligned, which will only bring negligible extra parameters. 
% However, their optimization objective is not specifically designed for searching a good network.
%
% From the perspective of architecture, some algorithms pre-define the pruned architecture~\cite{li2017pruning,he2017channel,he2018soft,liu2017learning}.
% Other algorithms implicitly find a data-specific architecture~\cite{wen2016learning,alvarez2016learning,he2018amc}, by automatically determining the pruning and compression ratio of each layer.
%We suggest that knowledge transfer is important to improve the accuracy of unpruned network.

Network transformation~\cite{chen2016net2net,gordon2018morphnet,cai2018efficient} also studied the depth and width of networks. Chen~et~al.~\cite{chen2016net2net} manually widen and deepen a network, and proposed Net2Net to initialize the lager network.
Ariel~et~al.~\cite{gordon2018morphnet} proposed a heuristic strategy to find a suitable width of networks by alternating between shrinking and expanding.
Cai~et~al.~\cite{cai2018efficient} utilized a RL agent to grow the depth and width of CNNs, while our {\NAME} is a differentiable approach and can not only enlarge but also shrink CNNs.
%proposed Net2Net to make a network deeper and wider with good

Knowledge transfer has been proven to be effective in the literature of pruning.
%Literature has shown that knowledge transfer has proven to be effective in pruning. 
The parameters of the networks can be transferred from the pre-trained initialization~\cite{li2017pruning,he2018soft}.
% Dedicated designs for the transfer mechanism are required, such as block-wise reconstruction loss~\cite{minnehan2019cascaded}.
Minnehan~et~al.~\cite{minnehan2019cascaded} transferred the knowledge of uncompressed network via a block-wise reconstruction loss.
In this paper, we apply a simple KD approach~\cite{hinton2014distilling} to perform knowledge transfer, which achieves robust performance for the searched architectures.

\section{Methodology}\label{sec:method}

Our pruning approach consists of three steps:
(1) training the unpruned large network by a standard classification training procedure.
(2) searching for the depth and width of a small network via the proposed {\NAME}.
(3) transferring the knowledge from the unpruned large network to the searched small network by a simple KD approach~\cite{hinton2014distilling}.
We will introduce the background, show the details of {\NAME}, and explain the knowledge transfer procedure.
%Note that we We name our pruning algorithm as DSTC, since its core idea is a differentiable search algorithm DSTC.

\begin{figure}
\centering
\includegraphics[width=\textwidth]{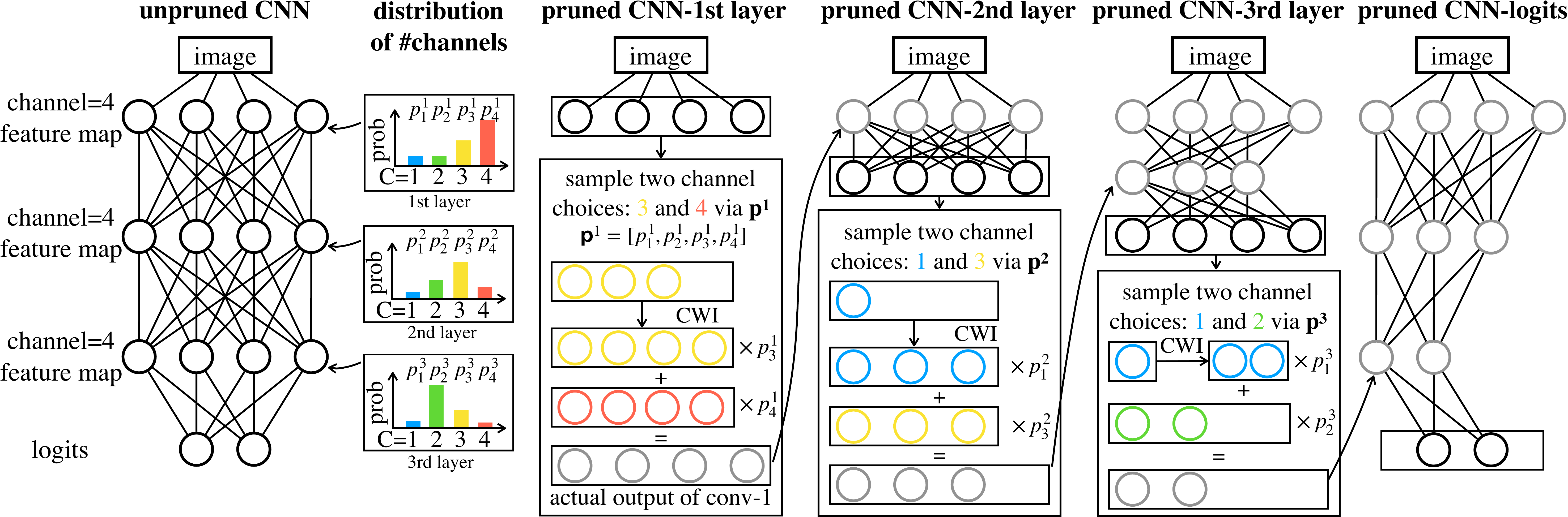}
\caption{
Searching for the width of a pruned CNN from an unpruned three-layer CNN.
% We take a CNN with three convolutional layers as an example for searching width. 
Each convolutional layer is equipped with a learnable distribution for the size of the channels in this layer, indicated by $\vp^{i}$ on the left side.
The feature map for every layer is built sequentially by the layers, as shown on the right side.
For a specific layer, K (2 in this example) feature maps of different sizes are sampled according to corresponding distribution and combined by channel-wise interpolation ($\textrm{CWI}$) and weighted sum.
This aggregated feature map is fed as input to the next layer.
% When we forward the first layer, we first sample two numbers based on $\vp^{1}$, e.g., \textcolor{yellow}{three channels} and \textcolor{red}{four channels}.
% Suppose $\mF\in\sR^{4\times{H}\times{W}}$ denotes the original output feature map of this layer, where $H$ and $W$ indicates the spatial height and width of this feature map. Then, three channels and four channels corresponds to $A = \mF_{1:3,:,:}$ and $B = \mF_{1:4,:,:}$, respectively.
% We aggregate these two features by (1) channel-wise interpolate $A$ to the dimension as $B$ and (2) weighted sum $A$ and $B$ according to the corresponding probabilities. This aggregated feature is fed as input to the next layer.
}
\label{fig:framework}
\end{figure}

\subsection{Transformable Architecture Search}\label{sec:method-TAS}
Network channel pruning aims to reduce the number of channels in each layer of a network.
Given an input image, a network takes it as input and produces the probability over each target class.
Suppose $\mX$ and $\mO$ are the input and output feature tensors of the $l$-th convolutional layer (we take 3-by-3 convolution as an example), this layer calculates the following procedure:
\begin{align}
    \mO_{j} = \sum\nolimits_{k=1}^{c_{in}} \mX_{k,:,:} * \mW_{j,k,:,:} \hspace{3mm}\textrm{where}\hspace{2mm} 1 \leq j \leq c_{out},
\end{align}
\noindent where $\mW\in\sR^{c_{out}\times{c_{in}}\times3\times3}$ indicates the convolutional kernel weight, $c_{in}$ is the input channel, and $c_{out}$ is the output channel.
$\mW_{j,k,:, :}$ corresponds to the $k$-th input channel and $j$-th output channel.
$*$ denotes the convolutional operation.
Channel pruning methods could reduce the number of $c_{out}$, and consequently, the $c_{in}$ in the next layer is also reduced.

\textbf{Search for width}. We use parameters $\alpha \in \sR^{{|\sC|}}$ to indicate the distribution of the possible number of channels in one layer, indicated by $\sC$ and $\max(\sC) \leq c_{out}$.
% Specifically, we use $\alpha \in \sR^{{|\sC|}}$ to parameterize the distribution of the number of output channels in one layer, 
% where $\sC$ indicates the set for the possible number of channels for this layer, and 
% $|\sC|$ indicates the cardinality of $\sC$.
The probability of choosing the $j$-th candidate for the number of channels can be formulated as:
\begin{align}\label{eq:softmax}
    p_{j} = \frac{\exp(\alpha_{j})}{\sum\nolimits_{k=1}^{|\sC|}\exp(\alpha_{k})} \hspace{3mm} \textrm{where}\hspace{2mm} 1 \leq j \leq |\sC|,
\end{align}

% To search the candidate number of channels, we sample the number of channels in each layer based on $p_{j}$, and then construct a pruned network via the sampled numbers of channels.
%the validation loss of
% Next, we need to optimize the probability $p_{j}$ based on this sampled network so as to allow that the high-performance and low-cost network has a high probability of being sampled.
However, the sampling operation in the above procedure is non-differentiable, which prevents us from back-propagating gradients through $p_{j}$ to $\alpha_{j}$.
Motivated by~\cite{dong2019search}, we apply Gumbel-Softmax~\cite{jang2017categorical,maddison2017concrete} to soften the sampling procedure to optimize $\alpha$:
% Specifically, we first use Gumbel-Softmax on $\alpha$ as \Eqref{eq:gumbel} and then aggregate the feature map fragments of different sizes as \Eqref{eq:aggregate}, which can be formulated as follows:
\begin{align}
    \hat{p}_{j} = \frac{\exp((\log{(p_{j})} + \vo_{j})/\tau)}{\sum\nolimits_{k=1}^{|\sC|}\exp((\log{(p_{k})}+\vo_{k})/\tau)} \hspace{3mm} \textrm{s.t.}\hspace{2mm} \vo_{j} = -\log(-\log(u)) ~\hspace{1mm}\&\hspace{1mm}~ u \sim \gU(0,1), \label{eq:gumbel}
%    \hat{p}_{j} = \frac{\exp((\alpha_{j} + \vo_{j})/\tau)}{\sum\nolimits_{k=1}^{|\sC|}\exp((\alpha_{k}+\vo_{k})/\tau)} \hspace{3mm} \textrm{s.t.}\hspace{2mm} \vo_{j} = -\log(-\log(u)) ~\hspace{1mm}\&\hspace{1mm}~ u \sim \gU(0,1), \label{eq:gumbel}
\end{align}
where $\gU(0,1)$ means the uniform distribution between 0 and 1.
$\tau$ is the softmax temperature.
When $\tau\rightarrow0$, $\hat{p}=[\hat{p}_{1},...,\hat{p}_{j},...]$ becomes one-shot, and the Gumbel-softmax distribution drawn from $\hat{p}$ becomes identical to the categorical distribution. When $\tau\rightarrow\infty$, the Gumbel-softmax distribution becomes a uniform distribution over $\sC$.
The feature map in our method is defined as the weighted sum of the original feature map fragments with different sizes, where weights are $\hat{p}$. Feature maps with different sizes are aligned by channel wise interpolation (CWI) so as for the operation of weighted sum.
To reduce the memory costs, we select a small subset with indexes $\sI\subseteq[|\sC|]$ for aggregation instead of using all candidates.
Additionally, the weights are re-normalized based on the probability of the selected sizes, which is formulated as:
% After compute the sampling probability $\hat{p}=\{\hat{p}_{j} | 1\leq{j}\leq|\sC|\}$, we randomly sample several candidate channels $\sC_{\sI}$ for aggregation, where
% This aggregation procedure can be formulated as:
\begin{align}
    \hat{\mO} = \sum\nolimits_{j\in\sI} \frac{\exp((\log(p_{j}) + \vo_{j})/\tau)}{\sum\nolimits_{k\in\sI}\exp((\log(p_{k})+\vo_{k})/\tau)} \times \textrm{CWI}( {\mO}_{1:\sC_{j},:,:}, \max(\sC_{\sI}) ) \hspace{2mm}\textrm{s.t.}\hspace{2mm} \sI \sim \gT_{\hat{p}}, \label{eq:aggregate}
%    \hat{\mO} = \sum\nolimits_{j\in\sI} \frac{\exp((\alpha_{j} + \vo_{j})/\tau)}{\sum\nolimits_{k\in\sI}\exp((\alpha_{k}+\vo_{k})/\tau)} \times \textrm{CWI}( {\mO}_{1:\sC_{j},:,:}, \max(\sC_{\sI}) ) \hspace{2mm}\textrm{s.t.}\hspace{2mm} \sI \sim \gT_{\hat{p}}, \label{eq:aggregate}
\end{align}
where $\gT_{\hat{p}}$ indicates the multinomial probability distribution parameterized by $\hat{p}$.
%and  indicates the sampled indexes of $\sC$.
% $\textrm{CWI}$ indicates the channel-wise interpolation operation, which can be used to align feature map fragment to be the maximum channels selected in this layer.
% With the help of \Eqref{eq:aggregate}, we can back-propagate gradients through the sampling procedure to the parameters $\{\alpha\}$, which indicates the distribution of width candidates.
The proposed $\textrm{CWI}$ is a general operation to align feature maps with different sizes.
It can be implemented in many ways, such a 3D variant of spatial transformer network~\cite{jaderberg2015spatial} or adaptive pooling operation~\cite{he2015spatial}.
In this paper, we choose the 3D adaptive average pooling operation~\cite{he2015spatial} as $\textrm{CWI}$\footnote{The formulation of the selected $\textrm{CWI}$: suppose $\mB=\textrm{CWI}(\mA, C_{out})$, where $\mB\in{\sR}^{C_{out}HW}$ and $\mA\in{\sR}^{CHW}$; then $\mB_{i,h,w}=\textrm{mean}(\mA_{s:e-1,h,w})$, where $s=\lfloor\frac{i{\times}C}{C_{out}}\rfloor$ and $e=\lceil\frac{(i+1){\times}C}{C_{out}}\rceil$.
We tried other forms of $\textrm{CWI}$, e.g., bilinear and trilinear interpolation. They obtain similar accuracy but are much slower than our choice.}
, because it brings no extra parameters and negligible extra costs.
We use Batch Normalization~\cite{ioffe2015batch} before $\textrm{CWI}$ to normalize different fragments.
\Figref{fig:framework} illustrates the above procedure by taking $|\sI|=2$ as an example.

\textit{Discussion w.r.t. the sampling strategy in \Eqref{eq:aggregate}}.
This strategy aims to largely reduce the memory cost and training time to an acceptable amount by only back-propagating gradients of the sampled architectures instead of all architectures. Compared to sampling via a uniform distribution, the applied sampling method (sampling based on probability) could weaken the gradients difference caused by per-iteration sampling after multiple iterations.
%Besides, we also compare different number of selected channels in 

\textbf{Search for depth}. We use parameters $\beta \in \sR^{L}$ to indicate the distribution of the possible number of layers in a network with $L$ convolutional layers.
We utilize a similar strategy to sample the number of layers following \Eqref{eq:gumbel} and allow $\beta$ to be differentiable as that of $\alpha$, using the sampling distribution $\hat{q}_{l}$ for the depth $l$.
% We first compute $\{\hat{q}_{i} | 1\leq{i}\leq{L}\}$ from $\beta$ using the same way as \Eqref{eq:gumbel}. 
% The $\hat{q}_{l}$ indicates the sampling probability for the depth of $l$. 
We then calculate the final output feature of the pruned networks as an aggregation from all possible depths, which can be formulated as:
\begin{align}\label{eq:depth}
    %\mO_{out} = \sum\nolimits_{l=1}^{L} \hat{q}_{l} \times \textrm{CWI}(\hat{\mO}^{l}, \max(\sC_{\sI}^{1} \cup ... \cup \sC_{\sI}^{L})) ,
    \mO_{out} = \sum\nolimits_{l=1}^{L} \hat{q}_{l} \times \textrm{CWI}(\hat{\mO}^{l}, C_{out}) ,
\end{align}
where $\hat{\mO}^{l}$ indicates the output feature map via \Eqref{eq:aggregate} at the $l$-th layer. $C_{out}$ indicates the maximum sampled channel among all $\hat{\mO}^{l}$.
The final output feature map $\mO_{out}$ is fed into the last classification layer to make predictions. In this way, we can back-propagate gradients to both width parameters $\alpha$ and depth parameters $\beta$.
%Both of them construct the overall architecture parameters.
%denoted as $\{\alpha,\beta\}$.
%These parameters are optimized during search, and are used to derive the final searched architecture by selecting the candidate with the learned maximum probability.

\textbf{Searching objectives}.
% Formally, we denote a network as $\gA$ with weights $\omega_{\gA}$. 
The final architecture $\gA$ is derived by selecting the candidate with the maximum probability, learned by the architecture parameters $\sA$, consisting of $\alpha$ for each layers and $\beta$.
The goal of our {\NAME} is to find an architecture $\gA$ with the minimum validation loss $\gL_{val}$ after trained by minimizing the training loss $\gL_{train}$ as:
\begin{align}
    \min_{\gA} \gL_{val}(\omega^{*}_{\gA}, \gA)~\hspace{3mm}\textrm{s.t.}\hspace{3mm} \omega^{*}_{\gA}=\arg\min_{\omega} \gL_{train}(\omega, \gA) ,
\end{align}
where $\omega^{*}_{\gA}$ indicates the optimized weights of $\gA$. The training loss is the cross-entropy classification loss of the networks.
% \begin{align}
%     \gL_{train} = -\log(\frac{\exp(\vz_{y})}{\sum_{j=1}^{|\vz|}\exp(\vz_{j})}); \label{eq:loss-train}
% \end{align}
Prevailing NAS methods~\cite{liu2019darts,zoph2017neural,dong2019search,cai2018proxylessnas,real2019regularized} optimize $\gA$ over network candidates with different typologies, while our {\NAME} searches over candidates with the same typology structure as well as smaller width and depth. As a result, the validation loss in our search procedure includes not only the classification validation loss but also the penalty for the computation cost:
\begin{align}
    \gL_{val} = -\log(\frac{\exp(\vz_{y})}{\sum_{j=1}^{|\vz|}\exp(\vz_{j})}) + \lambda_{cost} \gL_{cost} ,\label{eq:loss-val}
\end{align}
where $\vz$ is a vector denoting the output logits from the pruned networks,
$y$ indicates the ground truth class of a corresponding input,
and $\lambda_{cost}$ is the weight of $\gL_{cost}$.
% We target on two objectives, i.e., high performance and low cost. To maximize the accuracy, we use the standard softmax with cross-entropy loss.
The cost loss encourages the computation cost of the network (e.g., FLOP) to converge to a target $R$ so that the cost can be dynamically adjusted by setting different $R$.
%The piece-wise computation cost loss is defined as:
We used a piece-wise computation cost loss as:
\begin{align}
    \gL_{cost} = \left\{\begin{matrix}
 \log(\sE_{cost}(\sA))  & F_{cost}(\sA) > (1+t)\times{R}\\ 
 0 & (1-t)\times{R} < F_{cost}(\sA) < (1+t)\times{R} \\ 
 -\log(\sE_{cost}(\sA)) & F_{cost}(\sA) < (1-t)\times{R}
\end{matrix}\right.,
\end{align} 
where $\sE_{cost}(\sA)$ computes the expectation of the computation cost, based on the architecture parameters $\sA$.
Specifically, it is the weighted sum of computation costs for all candidate networks, where the weight is the sampling probability.
$F_{cost}(\sA)$ indicates the actual cost of the searched architecture, whose width and depth are derived from $\sA$.
$t\in[0,1]$ denotes a toleration ratio, which slows down
\begin{wrapfigure}{r}{0.42\textwidth}
\begin{minipage}{0.42\textwidth}
  \vspace{-8mm}
  \begin{algorithm}[H]
  \caption{The {\NAME} Procedure}
  \label{alg:TAS}
  \begin{algorithmic}[1]
    \REQUIRE split the training set into two disjoint sets: $\sD_{train}$ and $\sD_{val}$
    \WHILE{not converge}
    	  \STATE Sample batch data $\sD_{t}$ from $\sD_{train}$
    	  \STATE Calculate $\gL_{train}$ on $\sD_{t}$ to update network weights
    	  \STATE Sample batch data $\sD_{v}$ from $\sD_{val}$
    	  \STATE Calculate $\gL_{val}$ on $\sD_{v}$ via \Eqref{eq:loss-val} to update $\sA$
    \ENDWHILE
    \STATE Derive the searched network from $\sA$
    \STATE Randomly initialize the searched network and optimize it by KD via \Eqref{eq:KD} on the training set
  \end{algorithmic}
  \end{algorithm}
  \vspace{-12mm}
\end{minipage}
\end{wrapfigure}
the speed of changing the searched architecture.
Note that we use FLOP to evaluate the computation cost of a network, and it is readily to replace FLOP with other metric, such as latency~\cite{cai2018proxylessnas}.

%\textbf{{\NAME}}.
We show the overall algorithm in \Algref{alg:TAS}.
During searching, we forward the network using \Eqref{eq:depth} to make both weights and architecture parameters differentiable.
%We first calculate $\gL_{train}$ on the training set to optimize the pruned networks' weights, and then calculate $\gL_{val}$ on the validation set to optimize the architecture parameters $\sA$.
We alternatively minimize $\gL_{train}$ on the training set to optimize the pruned networks' weights and $\gL_{val}$ on the validation set to optimize the architecture parameters $\sA$.
%These two steps are alternatively performed.
After searching, we pick up the number of channels with the maximum probability as width and the number of layers with the maximum probability as depth.
The final searched network is constructed by the selected width and depth.
This network will be optimized via KD, and we will introduced the details in \Secref{sec:method-KD}.

\subsection{Knowledge Transfer}\label{sec:method-KD}

Knowledge transfer is important to learn a robust pruned network, and we employ a simple KD algorithm~\cite{hinton2014distilling} on a searched network architecture.
This algorithm encourages the predictions $\vz$ of the small network to match soft targets from the unpruned network via the following objective:
\begin{align}
    \gL_{\textrm{match}} = - \sum\nolimits_{i=1}^{|\vz|} \frac{\exp(\hat{\vz}_{i}/T)}{\sum_{j=1}^{|\vz|}\exp(\hat{\vz}_{j}/T)} \log(\frac{\exp({\vz}_{i}/T)}{\sum_{j=1}^{|\vz|}\exp({\vz}_{j}/T)}) ,
\end{align}
where $T$ is a temperature, and $\hat{\vz}$ indicates the logit output vector from the pre-trained unpruned network.
%With a higher temperature, this objective $\gL_{\textrm{match}}$ makes the network pay more 
Additionally, it uses a softmax with cross-entropy loss to encourage the small network to predict the true targets. The final objective of KD is as follows:
\begin{align}\label{eq:KD}
    \gL_{\textrm{KD}} = -\lambda \log(\frac{\exp(\vz_{y})}{\sum_{j=1}^{|\vz|}\exp(\vz_{j})}) + (1-\lambda) \gL_{\textrm{match}} \hspace{3mm}\textrm{s.t.}\hspace{2mm} 0 \leq \lambda \leq 1,
\end{align}
where $y$ indicates the true target class of a corresponding input.
$\lambda$ is the weight of loss to balance the standard classification loss and soft matching loss.
After we obtain the searched network (\Secref{sec:method-TAS}), we first pre-train the unpruned network and then optimize the searched network by transferring from the unpruned network via \Eqref{eq:KD}.

\section{Experimental Analysis}\label{sec:exps}

We introduce the experimental setup in \Secref{sec:exp-data-set}.
We evaluate different aspects of {\NAME} in \Secref{sec:exps-ablation}, such as hyper-parameters, sampling strategies, different transferring methods, etc.
Lastly, we compare {\NAME} with other state-of-the-art pruning methods in \Secref{sec:exps-sota}.

\begin{figure}
     \centering
     \begin{subfigure}[t]{0.47\textwidth}
         \centering
         \includegraphics[width=\textwidth]{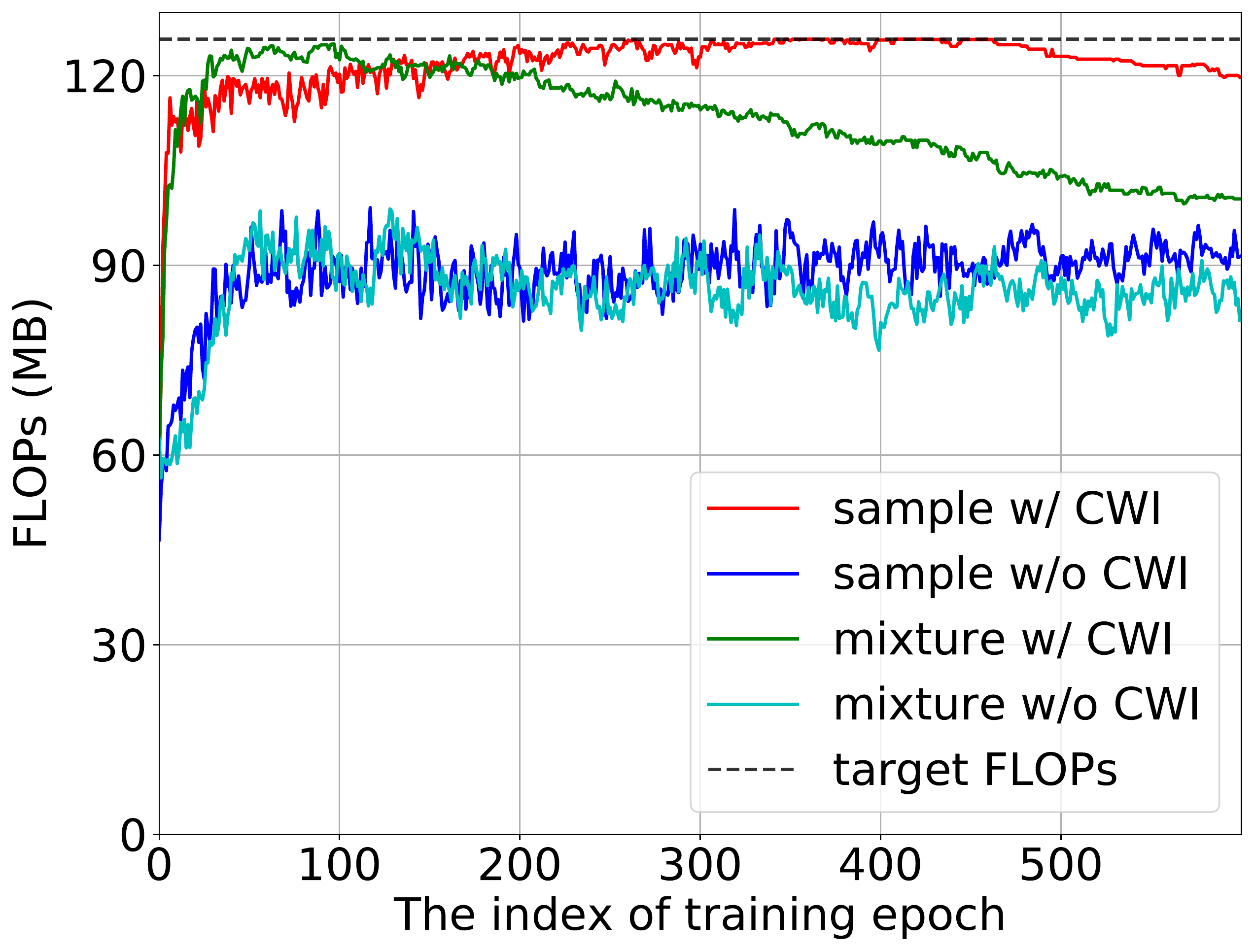}
         \caption{The FLOPs of the searched network over epochs when we do not constrain the FLOPs ($\lambda_{cost}=0$).}
         \label{fig:ablation-wrt-flop-00}
     \end{subfigure}
     \hfill
     \begin{subfigure}[t]{0.47\textwidth}
         \centering
         \includegraphics[width=\textwidth]{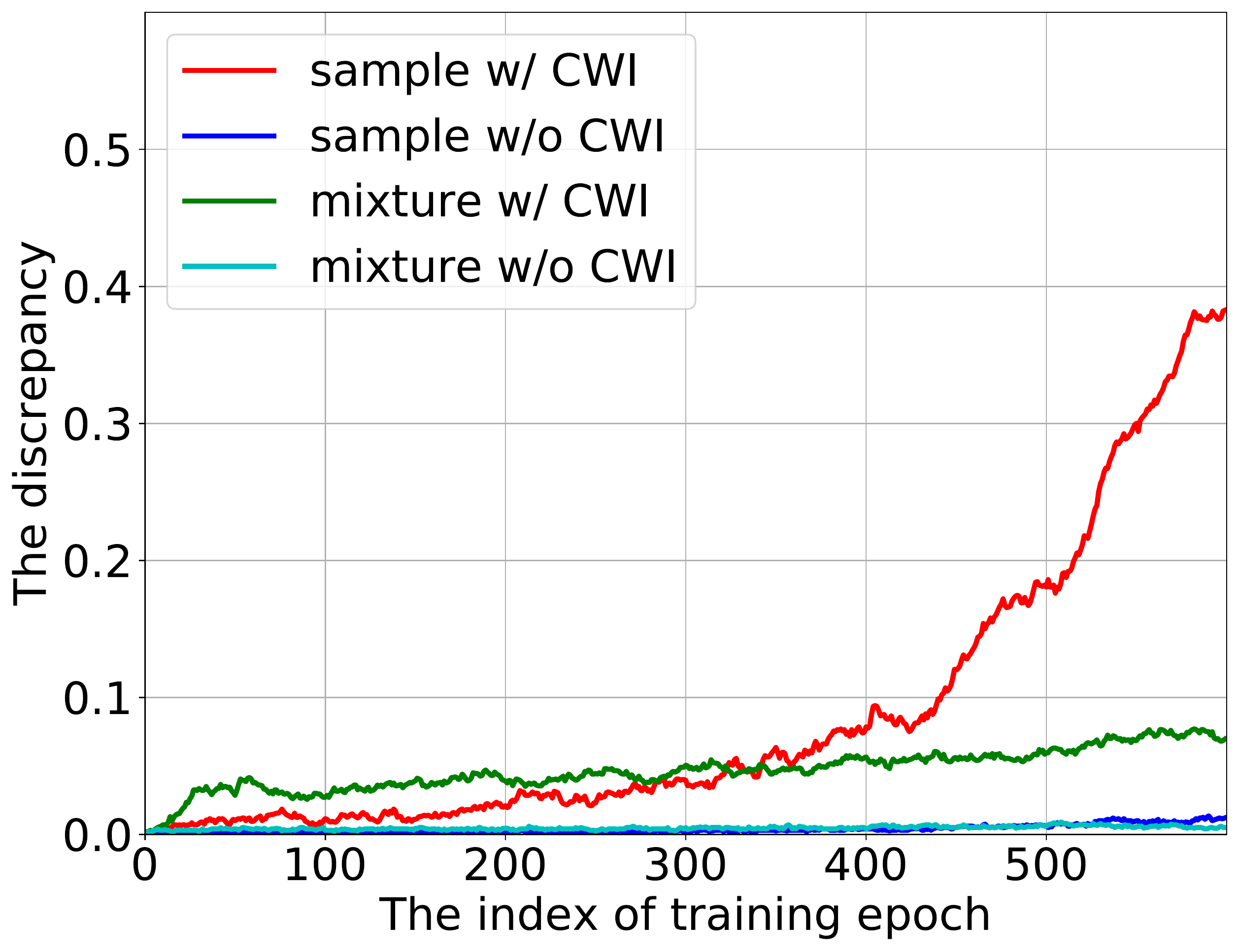}
         \caption{The mean discrepancy over epochs when we do not constrain the FLOPs ($\lambda_{cost}=0$).}
         \label{fig:ablation-wrt-dis-00}
     \end{subfigure}
     \begin{subfigure}[t]{0.47\textwidth}
         \centering
         \includegraphics[width=\textwidth]{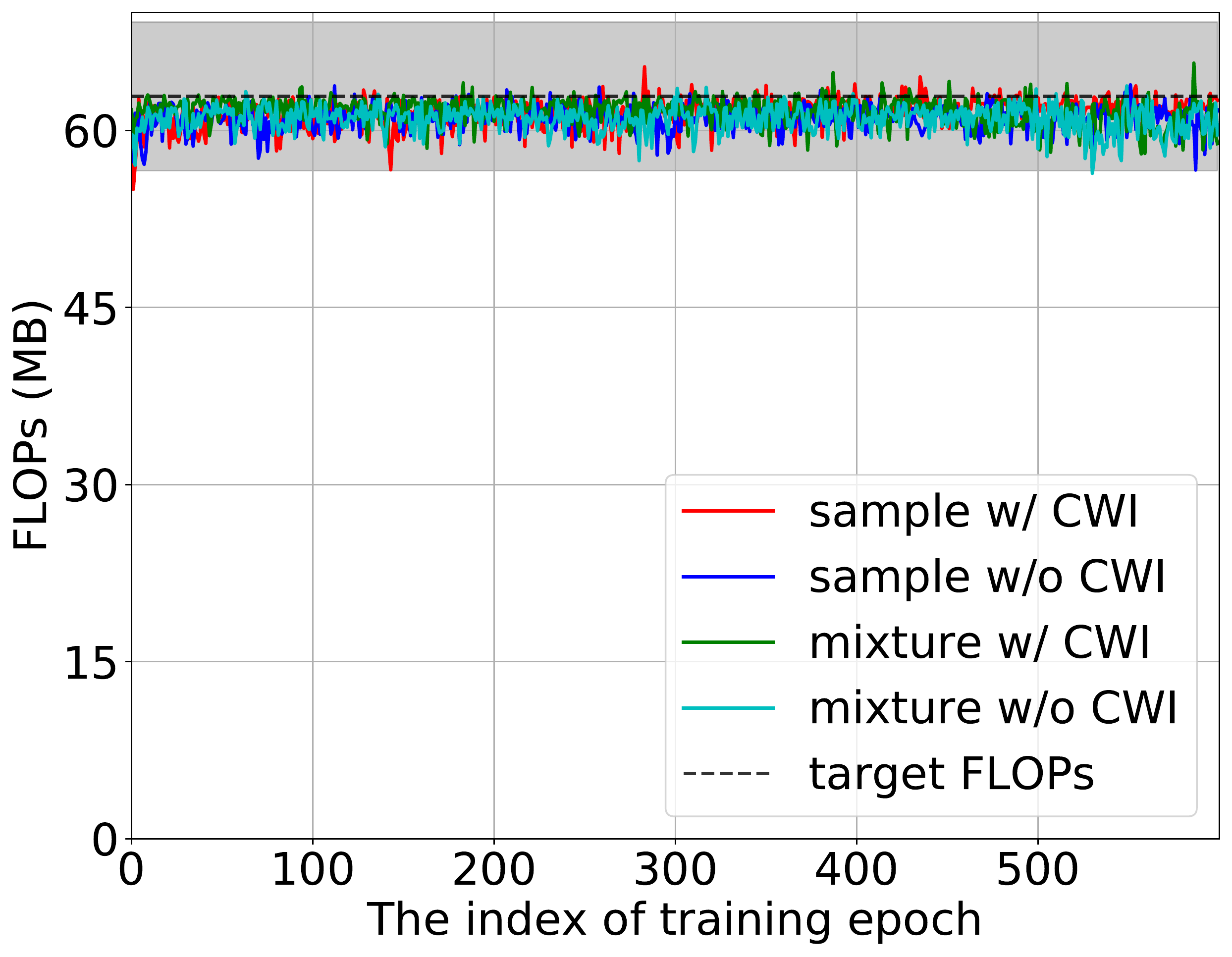}
         \caption{The FLOPs of the searched network over epochs when we constrain the FLOPs ($\lambda_{cost}=2$).}
         \label{fig:ablation-wrt-flop-05}
     \end{subfigure}
     \hfill
     \begin{subfigure}[t]{0.47\textwidth}
         \centering
         \includegraphics[width=\textwidth]{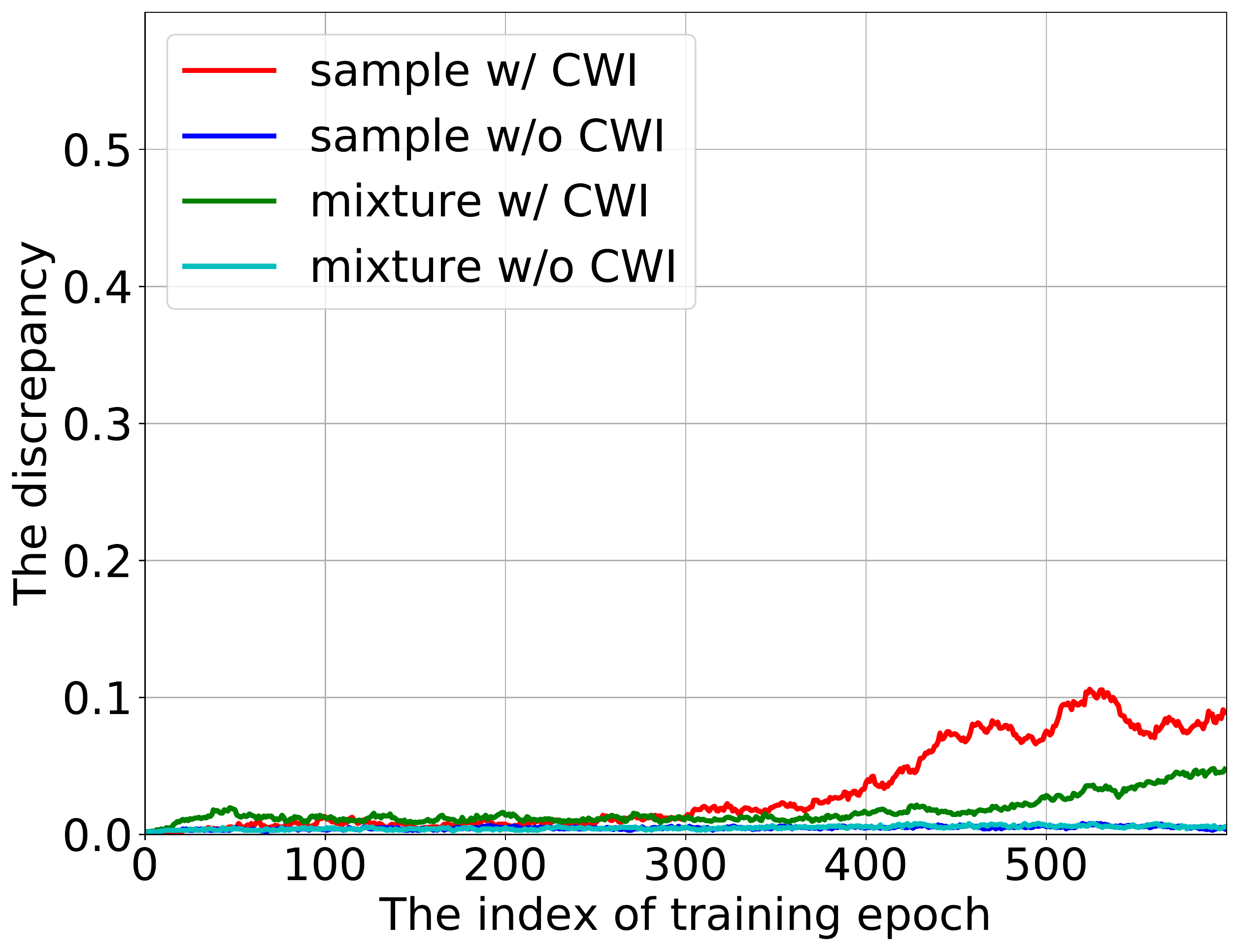}
         \caption{The mean discrepancy over epochs when we constrain the FLOPs ($\lambda_{cost}=2$).}
         \label{fig:ablation-wrt-dis-05}
     \end{subfigure}
        \caption{
        The impact of different choices to make architecture parameters differentiable.
        }
        \label{fig:ablation-4-cases}
\end{figure}

\subsection{Datasets and Settings}\label{sec:exp-data-set}

\textbf{Datasets.} We evaluate our approach on CIFAR-10, CIFAR-100~\cite{krizhevsky2009learning} and ImageNet~\cite{deng2009imagenet}. CIFAR-10 contains 50K training images and 10K test images with 10 classes. CIFAR-100 is similar to CIFAR-10 but has 100 classes. ImageNet contains 1.28 million training images and 50K test images with 1000 classes. We use the typical data augmentation of these three datasets.
On CIFAR-10 and CIFAR-100, we randomly crop 32$\times$32 patch with 4 pixels padding on each border, and we also apply the random horizontal flipping.
On ImageNet, we use the typical random resized crop, randomly changing the brightness / contrast / saturation, and randomly horizontal flipping for data augmentation.
During evaluation, we resize the image into 256$\times$256 and center crop a 224$\times$224 patch.

\textbf{The search setting.}
We search the number of channels over \{0.3, 0.4, 0.5, 0.6, 0.7, 0.8, 0.9, 1.0\} of the original number in the unpruned network. We search the depth within each convolutional stage.
We sample $|\sI|=2$ candidates in \Eqref{eq:aggregate} to reduce the GPU memory cost during searching.
We set $R$ according to the FLOPs of the compared pruning algorithms and set $\lambda_{cost}$ of 2.
We optimize the
\begin{wraptable}{r}{65mm}
\vspace{-3mm}
\caption{
The accuracy on CIFAR-100 when pruning about 40\% FLOPs of ResNet-32.}
\setlength{\tabcolsep}{3pt}
\begin{tabular}{l|c|c}
\hline
                          & FLOPs   & accuracy \\ \hline
Pre-defined               & 41.1 MB & 68.18 \% \\
Pre-defined w/ Init       & 41.1 MB & 69.34 \% \\
Pre-defined w/ KD         & 41.1 MB & 71.40 \% \\ \hline
Random Search             & 42.9 MB & 68.57 \% \\
Random Search w/ Init     & 42.9 MB & 69.14 \% \\
Random Search w/ KD       & 42.9 MB & 71.71 \% \\ \hline
{\NAME}$\dagger$          & 42.5 MB & 68.95 \% \\
{\NAME}$\dagger$ w/ Init  & 42.5 MB & 69.70 \% \\
{\NAME}$\dagger$ w/ KD ({\NAME}) & 42.5 MB & \textbf{72.41 \%} \\ 
\hline
\end{tabular}
\vspace{-4mm}
\label{table:alo-setting}
\end{wraptable}
weights via SGD and the architecture parameters via Adam.
For the weights, we start the learning rate from 0.1 and reduce it by the cosine scheduler~\cite{loshchilov2017sgdr}. For the architecture parameters, we use the constant learning rate of 0.001 and a weight decay of 0.001. On both CIFAR-10 and CIFAR-100, we train the model for 600 epochs with the batch size of 256.
On ImageNet, we train ResNets~\cite{he2016deep} for 120 epochs with the batch size of 256.
%On ImageNet, we train ResNets~\cite{he2016deep} for 120 epochs and MobileNet-V2~\cite{sandler2018mobilenetv2} for 150 epochs with the batch size of 256.
The toleration ratio $t$ is always set as 5\%.
The $\tau$ in \Eqref{eq:gumbel} is linearly decayed from 10 to 0.1.

\textbf{Training.} For CIFAR experiments, we use SGD with a momentum of 0.9 and a weight decay of 0.0005. We train each model by 300 epochs, start the learning rate at 0.1, and reduce it by the cosine scheduler~\cite{loshchilov2017sgdr}. We use the batch size of 256 and 2 GPUs.
When using KD on CIFAR, we use $\lambda$ of 0.9 and the temperature $T$ of 4 following~\cite{zagoruyko2017paying}.
For ResNet models on ImageNet, we follow most hyper-parameters as CIFAR, but use a weight decay of 0.0001. We use 4 GPUs to train the model by 120 epochs with the batch size of 256.
%For MobileNet-V2~\cite{sandler2018mobilenetv2}, we use SGD with the momentum of 0.9 and the weight decay of 0.00004. We train each model by 150 epochs, start the learning rate of 0.05, and reduce it by the cosine scheduler. We also set the dropout ratio to 0 in MobileNet-V2. We find this training strategy can make MobileNet-V2 converge faster and obtain a similar performance compared to that of~\cite{sandler2018mobilenetv2}. For all ImageNet experiments, we use the batch size of 256 and 4 GPUs.
When using KD on ImageNet, we set $\lambda$ as 0.5 and $T$ as 4 on ImageNet.
%and use the label smoothing with a value of 0.1.

\subsection{Case Studies}\label{sec:exps-ablation}
In this section, we evaluate different aspects of our proposed {\NAME}. We also compare it with different searching algorithm and knowledge transfer method to demonstrate the effectiveness of {\NAME}.
%first evaluate the impact of different strategies to make the architecture parameters differentiable. Next, we show the effectiveness of the two components in {\NAME} by comparing (1) the network structures generated in different ways and (2) different knowledge transferring strategies.

%compare the network structure generated from (1) pre-fined methods, (2) a simple NAS baseline~\cite{liu2019darts}, and (3) our {\NAME}. 

\textbf{The effect of different strategies to differentiate $\alpha$}.
We apply our {\NAME} on CIFAR-100 to prune ResNet-56.
We try two different aggregation methods, i.e., using our proposed $\textrm{CWI}$ to align feature maps or not. We also try two different kinds of aggregation weights, i.e., Gumbel-softmax sampling as \Eqref{eq:gumbel} (denoted as ``sample'' in \Figref{fig:ablation-4-cases}) and vanilla-softmax as \Eqref{eq:softmax} (denoted as ``mixture'' in \Figref{fig:ablation-4-cases}). Therefore, there are four different strategies, i.e., with/without $\textrm{CWI}$ combining with Gumbel-softmax/vanilla-softmax.
Suppose we do not constrain the computational cost, then the architecture parameters should be optimized to find the maximum width and depth. This is because such network will have the maximum capacity and result in the best performance on CIFAR-100.
We try all four strategies with and without using the constraint of computational cost. We show the results in
\begin{wraptable}{r}{85mm}
\vspace{-3mm}
\caption{
{Results of different configurations when prune ResNet-32 on CIFAR-10 with one V100 GPU.
``\#SC'' indicates the number of selected channels.
``H'' indicates hours.}
}
\setlength{\tabcolsep}{1pt}
\begin{tabular}{|l|c|c|c|c|c|}\hline
\makecell{\#SC} &  Search Time & Memory  & Train Time & FLOPs & Accuracy  \\ \hline
$|\sI|$=1       &  2.83 H      & 1.5GB       &  0.71 H    &  23.59 MB  &  89.85\%  \\ \hline
$|\sI|$=2       &  3.83 H      & 2.4GB       &  0.84 H    &  38.95 MB  &  92.98\%  \\ \hline
$|\sI|$=3       &  4.94 H      & 3.4GB       &  0.67 H    &  39.04 MB  &  92.63\%  \\ \hline
$|\sI|$=5       &  7.18 H      & 5.1GB       &  0.60 H    &  37.08 MB  &  93.18\%  \\ \hline
$|\sI|$=8       &  10.64 H     & 7.3GB       &  0.81 H    &  38.28 MB  &  92.65\%  \\ \hline
\end{tabular}
\vspace{-5mm}
\label{table:ablation-time}
\end{wraptable}
\Figref{fig:ablation-wrt-flop-05} and \Figref{fig:ablation-wrt-flop-00}.
When we do not constrain the FLOPs, our {\NAME} can successfully find the best architecture should have a maximum width and depth. However, other three strategies failed.
When we use the FLOP constraint, we can successfully constrain the computational cost in the target range.
We also investigate discrepancy between the highest probability and the second highest probability in \Figref{fig:ablation-wrt-dis-05} and \Figref{fig:ablation-wrt-dis-00}. Theoretically, a higher discrepancy indicates that the model is more confident to select a certain width, while a lower discrepancy means that the model is confused and does not know which candidate to select.
As shown in \Figref{fig:ablation-wrt-dis-05}, with the training procedure going, our {\NAME} becomes more confident to select the suitable width. In contrast, strategies without $\textrm{CWI}$ can not optimize the architecture parameters; and ``mixture with $\textrm{CWI}$'' shows a worse discrepancy than ours.

\textbf{Comparison w.r.t. structure generated by different methods} in \Tabref{table:alo-setting}.
%We compare three kinds of structures to prune ResNet-32 on CIFAR-100.
``Pre-defined'' means pruning a fixed ratio at each layer~\cite{li2017pruning}.
``Random Search'' indicates an NAS baseline used in~\cite{liu2019darts}.
%which first randomly samples 10 structures and fully trains them, and pick up the best one according to the validation accuracy.
``{\NAME$\dagger$}'' is our proposed differentiable searching algorithm.
We make two observations: (1) searching can find a better structure using different knowledge transfer methods; (2) our {\NAME} is superior to the NAS random baseline.

\begin{table}[t]
  \caption{
  Comparison of different pruning algorithms for ResNet on CIFAR.
  ``Acc'' = accuracy, ``FLOPs'' = FLOPs (pruning ratio), ``TAS (D)'' = searching for depth, ``TAS (W)'' = searching for width, ``TAS'' = searching for both width and depth.
  }
  \vspace{2mm}
  \centering
  \setlength{\tabcolsep}{2pt}
  \begin{tabular}{c c c c c c c c}
    \toprule
    % \multirow{2}{*}{FLOPs} & \multirow{2}{*}{\makecell{Prune\\Ratio}}
  \multirow{2}{*}{Depth}  & \multirow{2}{*}{Method} &       \multicolumn{3}{c}{CIFAR-10}       & \multicolumn{3}{c}{CIFAR-100}            \\
                          &                         & Prune Acc  & Acc Drop  &   FLOPs          & Prune Acc   & Acc Drop  &  FLOPs           \\
    \midrule
\multirow{6}{*}{20}
                          &LCCL~\cite{dong2017more} &  91.68\%   & 1.06\%    & 2.61E7 (36.0\%) &  64.66\%   &  2.87\%   & 2.73E7 (33.1\%)  \\
                          & SFP~\cite{he2018soft}   &  90.83\%   & 1.37\%    & 2.43E7 (42.2\%) &  64.37\%   &  3.25\%   & 2.43E7 (42.2\%)  \\
                          &FPGM~\cite{he2019pruning}&  91.09\%   & 1.11\%    & 2.43E7 (42.2\%) &  66.86\%   &  0.76\%   & 2.43E7 (42.2\%)  \\\cmidrule[0.5pt](lr){2-8}
                          & {\NAME} (D)             &  90.97\%   & 1.91\%    & 2.19E7 (46.2\%) &  64.81\%   &  3.88\%   & 2.19E7 (46.2\%)  \\
                          & {\NAME} (W)             &  92.31\%   & 0.57\%    & 1.99E7 (51.3\%) &  68.08\%   &  0.61\%   & 1.92E7 (52.9\%)  \\
                          & {\NAME}                 &  92.88\%   & 0.00\%    & 2.24E7 (45.0\%) &  68.90\%   &  -0.21\%  & 2.24E7 (45.0\%)  \\
    \midrule
\multirow{6}{*}{32}
                          &LCCL~\cite{dong2017more} &  90.74\%   & 1.59\%    & 4.76E7 (31.2\%) & 67.39\%    &  2.69\%   & 4.32E7 (37.5\%)  \\
                          & SFP~\cite{he2018soft}   &  92.08\%   & 0.55\%    & 4.03E7 (41.5\%) & 68.37\%    &  1.40\%   & 4.03E7 (41.5\%)  \\
                          &FPGM~\cite{he2019pruning}&  92.31\%   & 0.32\%    & 4.03E7 (41.5\%) & 68.52\%    &  1.25\%   & 4.03E7 (41.5\%)  \\\cmidrule[0.5pt](lr){2-8} %
                          & {\NAME} (D)             &  91.48\%   & 2.41\%    & 4.08E7 (41.0\%) & 66.94\%    &  3.66\%   & 4.08E7 (41.0\%)  \\
                          & {\NAME} (W)             &  92.92\%   & 0.96\%    & 3.78E7 (45.4\%) & 71.74\%    & -1.12\%   & 3.80E7 (45.0\%)  \\
                          & {\NAME}                 &  93.16\%   & 0.73\%    & 3.50E7 (49.4\%) & 72.41\%    & -1.80\%   & 4.25E7 (38.5\%)  \\
     \midrule
\multirow{7}{*}{56}
                          &PFEC~\cite{li2017pruning}&  93.06\%   & -0.02\%   & 9.09E7 (27.6\%) &   $-$      &   $-$     & $-$             \\
%                          & CP~\cite{he2017channel} &  91.80\%   & 1.00\%    & 6.29E7 (50.0\%) &   $-$      &   $-$     & $-$             \\
                          &LCCL~\cite{dong2017more} &  92.81\%   & 1.54\%    & 7.81E7 (37.9\%) &  68.37\%   &   2.96\%  &  7.63E7 (39.3\%) \\
                          & AMC~\cite{he2018amc}    &  91.90\%   & 0.90\%    & 6.29E7 (50.0\%) &    $-$     &   $-$     &  $-$            \\
                          & SFP~\cite{he2018soft}   &  93.35\%   & 0.56\%    & 5.94E7 (52.6\%) &  68.79\%   &   2.61\%  &  5.94E7 (52.6\%) \\
                          &FPGM~\cite{he2019pruning}&  93.49\%   & 0.42\%    & 5.94E7 (52.6\%) &  69.66\%   &   1.75\%  &  5.94E7 (52.6\%) \\\cmidrule[0.5pt](lr){2-8}
%                          &FPGM~\cite{he2019pruning}&  93.49\%   & 0.10\%    & 5.94E7 (52.6\%) &  69.66\%   &   1.75\%  &  5.94E7 (52.6\%) \\\cmidrule[0.5pt](lr){2-8}
                          & {\NAME}                 &  93.69\%   & 0.77\%    & 5.95E7 (52.7\%) &  72.25\%   &   0.93\%  &  6.12E7 (51.3\%)  \\
      \midrule
\multirow{5}{*}{110}
                          &LCCL\cite{dong2017more}  &  93.44\%   & 0.19\%    & 1.66E8 (34.2\%) &  70.78\%   &   2.01\%  &  1.73E8 (31.3\%)   \\
                          &PFEC~\cite{li2017pruning}&  93.30\%   & 0.20\%    & 1.55E8 (38.6\%) &   $-$      &    $-$    &    $-$  \\
                          &SFP~\cite{he2018soft}    &  92.97\%   & 0.70\%    & 1.21E8 (52.3\%) &  71.28\%   &   2.86\%  &  1.21E8 (52.3\%)   \\
                          &FPGM~\cite{he2019pruning}&  93.85\%   & -0.17\%    & 1.21E8 (52.3\%) &  72.55\%   &   1.59\%  &  1.21E8 (52.3\%)   \\\cmidrule[0.5pt](lr){2-8}
                          & {\NAME}                 &  94.33\%   & 0.64\%    & 1.19E8 (53.0\%) &  73.16\%   &   1.90\%  &  1.20E8 (52.6\%)  \\
      \midrule
\multirow{2}{*}{164}
                          &LCCL\cite{dong2017more}  &  94.09\%   & 0.45\%    & 1.79E8 (27.40\%)&  75.26\%   &   0.41\%  &  1.95E8 (21.3\%) \\\cmidrule[0.5pt](lr){2-8}
                          & {\NAME}                 &  94.00\%   & 1.47\%    & 1.78E8 (28.10\%)&  77.76\%   &   0.53\%  &  1.71E8 (30.9\%) \\
    \bottomrule
  \end{tabular}
  \vspace{-2mm}
  \label{table:CIFAR-SOTA}
\end{table}

\textbf{Comparison w.r.t. different knowledge transfer methods} in \Tabref{table:alo-setting}.
The first line in each block does not use any knowledge transfer method.
``w/ Init'' indicates using pre-trained unpruned network as initialization.
``w/ KD'' indicates using KD.
From \Tabref{table:alo-setting}, knowledge transfer methods can consistently improve the accuracy of pruned network, even if a simple method is applied (Init). Besides, KD is robust and improves the pruned network by more than 2\% accuracy on CIFAR-100.

\textbf{Searching width vs. searching depth.}
We try (1) only searching depth (``{\NAME} (D)''), (2) only searching width (``{\NAME} (W)''), and (3) searching both depth and width (``{\NAME}'') in \Tabref{table:CIFAR-SOTA}.
Results of only searching depth are worse than results of only searching width.
If we jointly search for both depth and width, we can achieve better accuracy with similar FLOP than both searching depth and searching width only.

\textbf{The effect of selecting different numbers of architecture samples $\sI$ in \Eqref{eq:aggregate}}.
We compare different numbers of selected channels in \Tabref{table:ablation-time} and did experiments on a single NVIDIA Tesla V100.
The searching time and the GPU memory usage will increase linearly to $|\sI|$.
When $|\sI|$=1, since the re-normalized probability in \Eqref{eq:aggregate} becomes a constant scalar of 1, the gradients of parameters $\alpha$ will become 0 and the searching failed.
When $|\sI|$>1, the performance for different $|\sI|$ is similar.

\textbf{The speedup gain.} As shown in \Tabref{table:ablation-time}, {\NAME} can finish the searching procedure of ResNet-32 in about 3.8 hours on a single V100 GPU . If we use evolution strategy (ES) or random searching methods, we need to train network with many different candidate configurations one by one and then evaluate them to find a best. In this way, much more computational costs compared to our {\NAME} are required.
A possible solution to accelerate ES or random searching methods is to share parameters of networks with different configurations~\cite{pham2018efficient,yu2019network}, which is beyond the scope of this paper.

\subsection{Compared to the state-of-the-art}\label{sec:exps-sota}

%We compare our {\NAME} with state-of-the-art algorithms on three benchmarks.
%: CIFAR-10, CIFAR-100, and ImageNet.

\textbf{Results on CIFAR} in \Tabref{table:CIFAR-SOTA}.
We prune different ResNets on both CIFAR-10 and CIFAR-100.
Most previous algorithms perform poorly on CIFAR-100, while our {\NAME} consistently outperforms then by more than 2\% accuracy in most cases.
On CIFAR-10, our {\NAME} outperforms the state-of-the-art algorithms on ResNet-20,32,56,110.
For example, {\NAME} obtains 72.25\% accuracy by pruning ResNet-56 on CIFAR-100, which is higher than 69.66\% of FPGM~\cite{he2019pruning}.
For pruning ResNet-32 on CIFAR-100, we obtain greater accuracy and less FLOP than the unpruned network.
We obtain a slightly worse performance than LCCL~\cite{dong2017more} on ResNet-164.
It because there are $8^{163}\times18^{3}$ candidate network structures to searching for pruning ResNet-164. It is challenging to search over such huge search space, and the very deep network has the over-fitting problem on CIFAR-10~\cite{he2016deep}.

\begin{table}
  \caption{
  Comparison of different pruning algorithms for different ResNets on ImageNet.
  }
  \centering
  \vspace{2mm}
  \setlength{\tabcolsep}{4pt}
  \begin{tabular}{c c c c c c c c}
    \toprule
  \multirow{2}{*}{Model}  & \multirow{2}{*}{Method} & \multicolumn{2}{c}{Top-1} & \multicolumn{2}{c}{Top-5} & \multirow{2}{*}{FLOPs} & \multirow{2}{*}{\makecell{Prune\\Ratio}} \\
                          &                         & Prune Acc    &  Acc Drop  & Prune Acc   &    Acc Drop &       &             \\
    \midrule
\multirow{4}{*}{ResNet-18}&LCCL~\cite{dong2017more} &  66.33\%     &  3.65\%    &  86.94\%     &   2.29\%   &  1.19E9 &   34.6\%   \\
                          & SFP~\cite{he2018soft}   &  67.10\%     &  3.18\%    &  87.78\%     &   1.85\%   &  1.06E9 &   41.8\%   \\
                          &FPGM~\cite{he2019pruning}&  68.41\%     &  1.87\%    &  88.48\%     &   1.15\%   &  1.06E9 &   41.8\%   \\\cmidrule[0.5pt](lr){2-8}
                          & {\NAME}                 &  69.15\%     &  1.50\%    &  89.19\%     &   0.68\%   &  1.21E9 &   33.3\% \\
%                          & {\NAME}                 & 67.93\%      &            &  88.58\%     &            &  1.06E9 &   41.8\%   \\
     \midrule
\multirow{6}{*}{ResNet-50}& SFP~\cite{he2018soft}   &  74.61\%     &  1.54\%    &  92.06\%     &   0.81\%   &  2.38E9 &   41.8\%   \\
                          & CP~\cite{he2017channel} &    -         &    -       &  90.80\%     &   1.40\%   &  2.04E9 &   50.0\%   \\
                          & Taylor~\cite{molchanov2019importance}&  74.50\%     &  1.68\%    &   -   &    - &  2.25E9 &   44.9\%   \\
                          & AutoSlim~\cite{yu2019network} &  76.00\%     &    -       &    -         &     -      &  3.00E9 &   26.6\%   \\
                          & FPGM~\cite{he2019pruning}& 75.50\%     &  0.65\%    &  92.63\%     &   0.21\%   &  2.36E9 &   42.2\%   \\\cmidrule[0.5pt](lr){2-8}
                          & {\NAME}                  & 76.20\%     &  1.26\%    &  93.07\%     &   0.48\%   &  2.31E9 &   43.5\%   \\
    \bottomrule
%    \midrule
%\multirow{3}{*}{ResNet-101}& SNLI~\cite{ye2018rethinking}& 75.27\% &  2.10\%    &  -           &     -      &  4.13E9 & 47.0\% \\  
%                           & FPGM~\cite{he2019pruning}& 77.32\%    &  0.05\%    &  93.56\%     &  0.00\%    &  4.51E9 & 42.2\% \\\cmidrule[0.5pt](lr){2-8}
%                           & {\NAME} \\
%     \midrule
%\multirow{2}{*}{MobileNet-V2} & AMC~\cite{he2018amc}&    70.80\%   &   1.00\%   &       -      &       -    & 1.50E8 & 50.0\% %\\\cmidrule[0.5pt](lr){2-8}
%                              & {\NAME}             &    70.90\%   &   1.18\%   &  89.74\%     &   0.76\%   & 1.49E8 & 50.3\% \\
%    \bottomrule
  \end{tabular}
  \vspace{-2mm}
  \label{table:ImageNet-SOTA}
\end{table}

\textbf{Results on ImageNet} in \Tabref{table:ImageNet-SOTA}.
We prune ResNet-18 and ResNet-50 on ImageNet.
For ResNet-18, it takes about 59 hours to search for the pruned network on 4 NVIDIA Tesla V100 GPUs.
The training time of unpruned ResNet-18 costs about 24 hours, and thus the searching time is acceptable.
With more machines and optimized implementation, we can finish {\NAME} with less time cost.
We show competitive results compared to other state-of-the-art pruning algorithms.
For example, {\NAME} prunes ResNet-50 by 43.5\% FLOPs, and the pruned network achieves 76.20\% accuracy, which is higher than FPGM by 0.7.
Similar improvements can be found when pruning ResNet-18.
%Similar improvements can be found when pruning ResNet-18 and MobileNet-V2.
Note that we directly apply the hyper-parameters on CIFAR-10 to prune models on ImageNet, and thus {\NAME} can potentially achieve a better result by carefully tuning parameters on ImageNet.

Our proposed {\NAME} is a preliminary work for the new network pruning pipeline. This pipeline can be improved by designing more effective searching algorithm and knowledge transfer method.
We hope that future work to explore these two components will yield powerful compact networks.

\section{Conclusion}\label{sec:conclusion}

In this paper, we propose a new paradigm for network pruning, which consists of two components.
For the first component, we propose to apply NAS to search for the best depth and width of a network. Since most previous NAS approaches focus on the network topology instead the network size, we name this new NAS scheme as Transformable Architecture Search (TAS).
Furthermore, we propose a differentiable TAS approach to efficiently and effectively find the most suitable depth and width of a network. For the second component, we propose to optimize the searched network by transferring knowledge from the unpruned network.
In this paper, we apply a simple KD algorithm to perform knowledge transfer, and conduct other transferring approaches to demonstrate the effectiveness of this component.
Our results show that new efforts focusing on searching and transferring may lead to new breakthroughs in network pruning.

\bibliographystyle{abbrv}{\small
\bibliography{abrv,ms}
}

\end{document}